# Exploring the Synergies of Hybrid CNNs and ViTs Architectures for Computer Vision: A survey

Haruna Yunusa[1], Shiyin Qin[1], Abdulrahman Hamman Adama Chukkol[2]

Abdulganiyu Abdu Yusuf[3], Isah Bello,[4] Adamu Lawan[5]

[1]School of Automation Science and Electrical Engineering, Beihang University, Beijing, China

[2]School of Information and Electronics, Beijing Institute of Technology, China

[3]School of Computer Science, Beijing Institute of Technology, China.

[4]School of Electrical and Information Engineering, Tianjin University, Tianjin, China

[5]School of Computer Science and Technology, Beihang University, Beijing, China

**Abstract**

The hybrid of Convolutional Neural Network (CNN) and Vision Transformers (ViT) architectures has emerged as a groundbreaking approach, pushing the boundaries of computer vision (CV). This comprehensive review provides a thorough examination of the literature on state-of-the-art hybrid CNN-ViT architectures, exploring the synergies between these two approaches. The main content of this survey includes: (1) a background on the vanilla CNN and ViT, (2) systematic review of various taxonomic hybrid designs to explore the synergy achieved through merging CNNs and ViTs models, (3) comparative analysis and application task-specific synergy between different hybrid architectures, (4) challenges and future directions for hybrid models, (5) lastly, the survey concludes with a summary of key findings and recommendations. Through this exploration of hybrid CV architectures, the survey aims to serve as a guiding resource, fostering a deeper understanding of the intricate dynamics between CNNs and ViTs and their collective impact on shaping the future of CV architectures.

## 1 Introduction

Computer vision (CV) is a rapidly evolving field with a wide range of applications, including image classification (Sheykhmousa et al., 2020; Chen et al., 2021; Azizi et al., 2021; Sun et al., 2019), object recognition (Zhao et al., 2020; Xie et al., 2020; Chen et al., 2019b), object tracking (Meinhardt et al., 2022; Kristan et al., 2019; Brasó & Leal-Taixé, 2020; Pang et al., 2021), semantic segmentation (Strudel et al., 2021; Zhou et al., 2022; Xie et al., 2021; Zhang et al., 2019; Zhong et al., 2020), and scene understanding (Hou et al., 2021; Muhammad et al., 2022; Roberts et al., 2021; Chen et al., 2019a; Sakaridis et al., 2021). Furthermore, convolutional neural networks (CNN) have been the dominant paradigm in CV since the introduction of AlexNet by Krizhevsky et al. (2012), in the early 2010s. Interestingly, this groundbreaking architecture marked a pivotal moment, catalyzing a shift towards deep learning (DL) and significantly influencing subsequent advancements in the field VGGNet (Simonyan & Zisserman, 2014b), GoogleNet (Szegedy et al., 2015), ResNet (He et al., 2016), DenseNet (Huang et al., 2017), MobileNet (Howard et al., 2017), NASNet (Zoph et al., 2018), Xception (Chollet, 2017), and EfficientNet (Tan & Le, 2019). Since then, CNN-based models have consistently achieved state-of-the-art (SOTA) results on the most prominent CV benchmarks such as IMAGENET (Deng et al., 2009), COCO (Lin et al., 2014), PASCAL VOC (Everingham et al., 2007), and MNIST (LeCun et al., 1998) thereby demonstrating their efficacy and becoming the de facto standard for CV tasks.

However, recent breakthroughs have challenged this dominance with the emergence of Vision in Transformers (ViT) by Dosovitskiy et al. (2020), inspiring subsequent designs like DeiT (Touvron et al., 2021), Swin (Liu et al., 2021), and T2T-ViT (Yuan et al., 2021). These designs have demonstrated their ability to outperform CNNs on certain tasks, especially those involving long-range dependencies in complex images. While both CNNs and ViTs exhibit unique strengths, harnessing the synergies between these architectures has become a focal point in advancing CV capabilities (Liu et al., 2021).

The recognition of the complementary nature of CNNs and ViTs has paved the way for exploring and developing hybrid architectures (Khan et al., 2023). Notably, these hybrid models aim to leverage the spatial hierarchies captured by CNNs and the attention mechanisms inherent in ViTs, merging these salient attributes to achieve enhanced performance across a spectrum of CV tasks. CNNs are particularly well-suited for learning local spatial features in images and are known for their efficiency in training and deployment, making them ideal for real-time applications. However, CNNs may face challenges in learning long-range dependencies, which can impact their performance on certain tasks like object detection (OD) and scene understanding. On the other hand, ViTs excel in learning long-range dependencies and capturing global context. Nevertheless, ViTs can be computationally expensive to train and deploy, rendering them less suitable for real-time applications

While the fusion of CNNs and ViTs in hybrid architectures holds promise for overcoming limitations and maximizing the strengths of each approach, it introduces challenges that must be addressed. One notable challenge is the increased complexity in designing and training hybrid architectures compared to traditional CNNs or ViTs. Achieving optimal performance demands a careful balance between the contributions of each component. Additionally, computational expenses pose another challenge, particularly when training and deploying large-scale hybrid architectures tailored for tackling complex CV tasks.

While previous surveys have focused on hybrids (Khan et al., 2023) or ViTs (Han et al., 2022), our survey delves into diverse design approaches that explores the synergies between CNN and ViT models. These approaches include parallel and serial integration, hierarchical integration, early and late feature fusion, as well as cross-attention calibration and MHSA integration. By bridging the gap between standalone CNN and ViT architectures, these hybrid designs elevate CV to new levels of accuracy, versatility, and robustness. This survey aims to provide researchers and developers with a comprehensive understanding of synergies in hybrid architectures, aiding in design decisions, performance evaluation, overcoming challenges, fostering innovation, and guiding practical applications. The subsequent sections of the paper follow the structure outlined in Fig. 1."

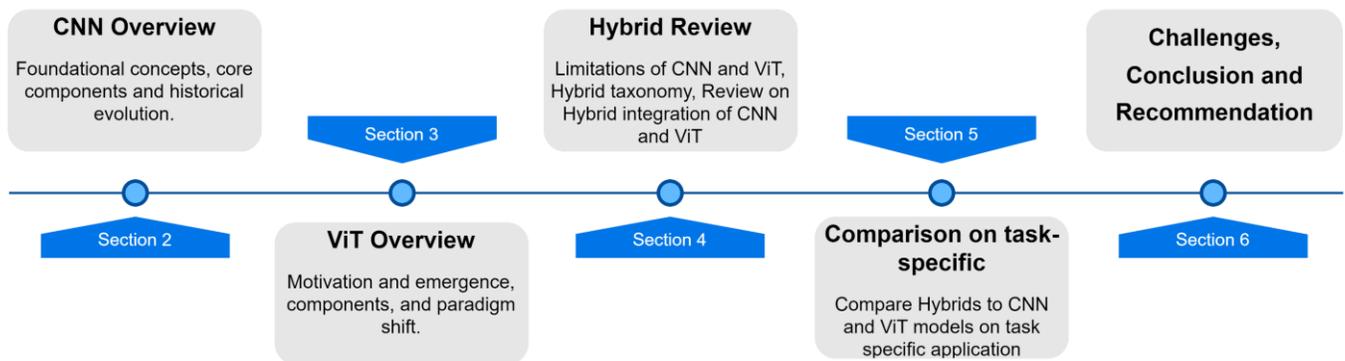

**Fig. 1** Research outline

## 2 CNN Overview

This section, provides an overview of CNN foundational concepts, architectural components, and historical evolution.

### 2.1 Foundational Concepts

CNN is a deep neural network that excels at image recognition and classification tasks. Inspired by the human visual cortex by Fukushima, (1980), they consist of several layers designed to detect specific patterns in the visual field, such as edges, textures, and colors. These patterns are then used to identify and classify objects within images (Fig. 2).

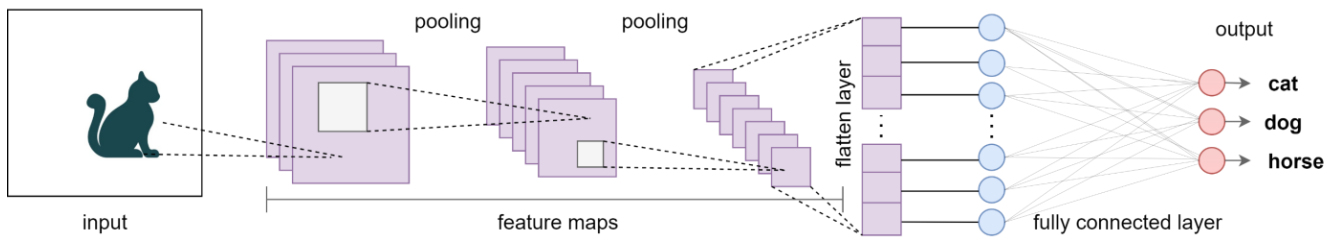

**Fig. 2** A typical CNN network

**2.1.1 Convolutional layers:** extract essential features from input images, operate by sliding a convolutional kernel across the image. This kernel, a small matrix of weights, performs a dot product with the corresponding image patch at each location, generating a feature map, a matrix indicating the feature strength at every image position (Fig. 3).

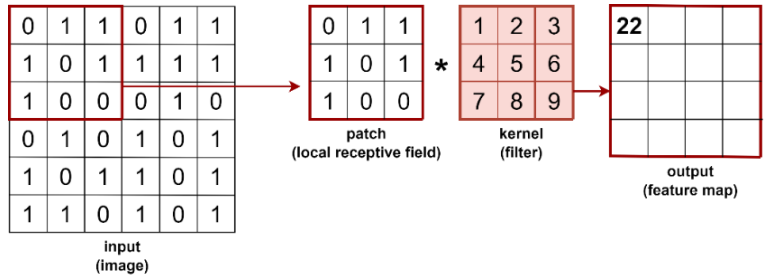

**Fig. 3** Convolutional operation

The formulation for a 2D convolution operation (Eqn. 1), considering an input matrix $I$ and a kernel matrix $K$, to compute the value at position $(i,j)$ in the output matrix $O$ is:

$$O(i,j) = \sum_{p=0}^{F-1}\sum_{q=0}^{F-1} I(i+p, j+q) \times K(p,q) \qquad \text{Eqn. (1)}$$

Eqn. 1 calculates the output value at position $(i,j)$ by taking the sum of element-wise products between the filter $K$ and the corresponding patch of the input matrix $I$ at that position.

**2.1.2 Pooling layer:** reduce feature map spatial dimensionality, introduce translation invariance, extract salient features, and prevent overfitting. Downsample feature maps by taking the maximum or average value of a small rectangular region (Fig. 4). This reduces the number of parameters in the network and makes it less computationally expensive to train.

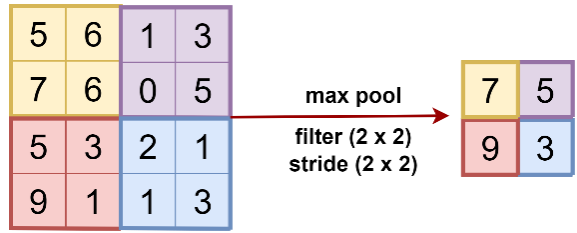

**Fig. 4** CNN pooling operation

For a max pooling operation (Eqn. 2), considering an input matrix $I$ and a pooling window size of $F$, to compute the value at position $(i,j)$ in the output matrix $O$ is:

$$(i,j) = max_{p=0}^{F-1} max_{q=0}^{F-1} I(i \times F + p, j \times F + q) \qquad \text{Eqn. (2)}$$

Eqn. 2 calculates the output value at position $(i,j)$ by taking the maximum value within the pooling window of size $(F \times F)$ at that position in the input matrix.

**2.1.3 Fully connected layer:** similar to fully connected layer in standard neural network. Each neuron is connected to every neuron in the previous layer, and the weights of these connections are learned during training. These layers classify images using functions like softmax (Fig. 5). The computation within this layer is represented in (Eqn. 3):

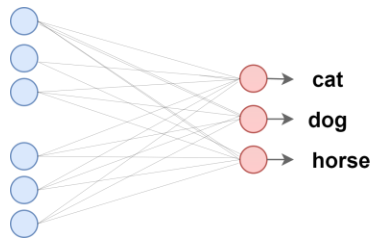

**Figure 5** CNN fully connected layer

$$y = Wx + b \qquad \text{Eqn. (3)}$$

Where, $W$ is the weights matrix, $x$ is the input vector to the layer, and $b$ is the bias vector added element-wise to the results of $Wx$ to obtain the final output vector y.

**2.2 Successful CNN components across various applications**

Over the decade, CNN models have been the most effective mechanism in CV tasks, consistently achieving SOTA results on the most prominent benchmarks, such as IMAGENET, COCO, PASCAL VOC, etc. This remarkable success can be attributed to a combination of key components that work together to enable CNNs to learn complex patterns from images, (Table 1).

Table 1 Key CNN components

| Components | Details |
| --- | --- |
| Multiple conv. layers | To extract more complex spatial features from the input image. |
| Different filter sizes | To extract features from various scales. |
| Dropout layers | To prevent overfitting by randomly dropping out neurons during training. |
| Batch normalization | To stabilize the training process and make it more efficient. |
| Transfer learning | To enhance the performance of CNNs on new tasks by leveraging pre-trained models from other related or different tasks. |
| Pooling layers | To reduce spatial dimensions, aiding in computational efficiency and feature translation invariance |
| Activation Functions | To introduce non-linearity, allowing the network to learn complex patterns. |
| Skip-connections | To enhance information flow across different layers, they alleviate the vanishing gradient problem, enabling the training of very deep networks. |
| Attention mechanism | To enable CNN model focus on specific parts of an input. |
| Data augmentation | Techniques like rotation, scaling, or flipping of input data augment the training dataset, enhancing model robustness and reducing overfitting. |

**2.3 CNN evolution and architectural diversity**

This section explores the evolution of CNN, tracing their historical development and listing key architectural advancements that have shaped the landscape of CV. We begin with the early foundations of CNN (Table 2).

Table 2 CNN prominent historical evolution

| Model | Details |
| --- | --- |
| Neocognitron (Fukushima,1980) | **Pioneering Visual Perception** developed by Kunihiko Fukushima, represents an early model foundational to modern CNNs. It mimics the visual processing mechanisms of the human brain, specifically for pattern recognition in images. Utilizing hierarchical layers of simple and complex cells, it introduced concepts like local receptive fields, weight sharing, and shift-invariance, laying the groundwork for subsequent CNN advancements. |
| LeNet-5 (LeCun et al., 1998) | **Pioneer of CNN** design for handwritten digit recognition, it introduces the core elements of modern CNNs, including convolutional, pooling, and fully connected layers. Its success in achieving SOTA results on the MNIST dataset set the stage for subsequent CNN advancements. |
| AlexNet (Krizhevsky et al., 2012) | **Unveiled DL potential** marks a turning point in DL. Designs for ImageNet benchmark, AlexNet featured deep convolutional layers, achieving a staggering 15% error rate compared to the previous best of 25%. This breakthrough demonstrated the potential of DL and paved the way for the current era of deep CNNs. |
| VGGNet (Simonyan & Zisserman 2014b) | **Simplicity and Depth** the Visual Geometry Group (VGG) at Oxford University proposed VGGNet, utilizes uniform architecture with small receptive fields and deep stacks of layers, demonstrating the effectiveness of network depth for feature learning. VGG16 and VGG19, became widely used benchmarks for evaluating CNN performance. |
| GoogLeNet (Szegedy et al., 2014) | **Efficiency and Performance** addresses the computational complexity of deep CNNs. It employs inception modules, which combine convolutional operations of different filter sizes in parallel, reducing computational cost while maintaining performance. This innovation enabled the training of deeper CNNs and improved model efficiency. |
| ResNet (He et al., 2015) | **Vanishing Gradient** revolutionizes training of deep CNNs through skip connections that bypass several layers, improving vanishing gradient problem that hinders training of deep networks. This breakthrough enables the training of extremely deep CNNs, achieving significant performance improvements. |
| DenseNet (G. Huang et al., 2016) | **Dense Connections** further enhances CNN architecture by proposing densely connected layers. It connects all layers directly, enabling each layer to receive feature maps from all preceding layers. This dense connectivity promotes feature reuse and improves parameter efficiency, leading to improve performance and reduced computational cost. |
| MobileNet (Howard & Zaremba, 2017) | **Lightweight Mobile Architectures** designs CNN for mobile and embedded vision applications. These architectures employ depth-wise separable convolutions to build lightweight and efficient models, suitable for resource-constrained environments without compromising performance. |
| EfficientNet (Sun et al. in 2019) | **Scalable and Efficient Networks** proposes variants of scalable and efficient CNN architectures by systematically scaling the network width, depth, and resolution. These models achieve SOTA performance with significantly fewer parameters and FLOPS, demonstrating superior efficiency across various tasks. |

## 2.4 Factors driving diverse CNN architecture design across applications

The diversity of CNN architectures stems from the need to address various challenges and exploit different task-specific characteristics (Khan et al., 2020), see Table 3 for details.

**Table 3** Key factors influencing CNN architectural design in CV.

| Factors | Details |
|---|---|
| Problem Complexity | The task complexity dictates architecture choice; simpler tasks like digit recognition may need fewer layers, while complex tasks like object detection or image segmentation require deeper architectures with advanced feature extraction (Simonyan & Zisserman, 2014a; He et al., 2017; Orsic et al., 2019). |
| Computational Resources | Computational resources availability influences architectural design. For resource-constrained devices like mobile phones, lightweight architectures like MobileNet are preferred, while powerful GPUs can accommodate deeper and complex architectures (Howard et al., 2017; Tan & Le, 2019; Iandola et al., 2016). |
| Data Availability | The unavailability of training data affects the choice of architecture. Regularization techniques, such as dropout and data augmentation, are often incorporated into the architecture (Srivastava et al., 2014; Engstrom et al., 2019; Zoph et al., 2020). |
| Task-Specific Requirements | Specific tasks have unique requirements that necessitate tailored architectures. For example, 3D object recognition requires architectures that processes volumetric data, while video analysis require architectures that captures temporal information (Maturana & Scherer, 2015; Simonyan & Zisserman, 2014a; Wang et al., 2018). |
| Architectural Innovations | Ongoing research continues introducing novel architectural designs, e.g. attention mechanisms, NAS techniques, and hybrid architectures that combine CNN with other DL models (Liu et al., 2018; Dosovitskiy et al., 2020). |

## 3 ViT overview

In this section, we discuss the emergence of ViT, motivation behind their design, and a comprehensive overview of their architectural components.

### 3.1 Motivation and Emergence

Since inception CNN has dominated the field of CV tasks, despite its success, its limitations in capturing global context and long-range dependencies have motivated a path-changing approach to developing CV architectures. Dosovitskiy et al. (2020) marks a paradigm shifts in CV through the introduction of ViT. Unlike CNN, which process images in a spatially localized manner, ViTs treat images as a sequence of tokens, much like natural language processing (NLP) tasks. This novel approach leverages the self-attention mechanism, a technique commonly used in NLP, to enable ViT capture global relationship between pixels in an image (Fig. 6).

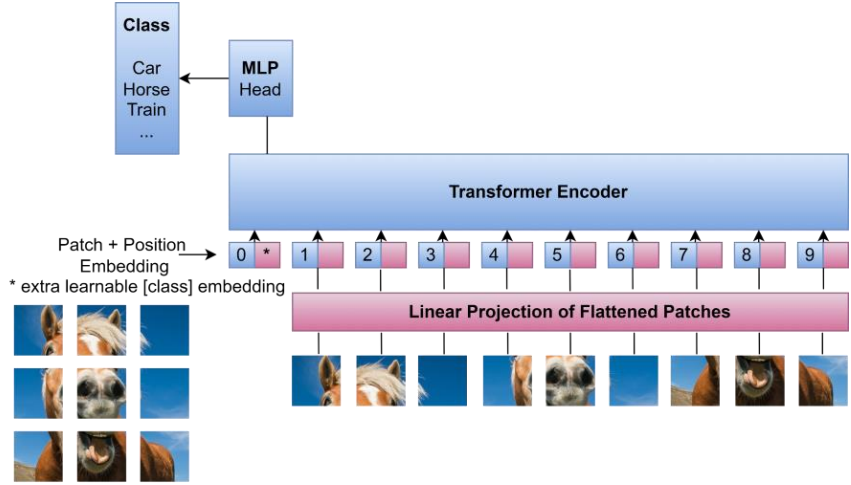

**Fig. 6** ViT architecture

## 3.2 ViT architectural components

The ***self-attention mechanism*** is the core component of ViT (Dosovitskiy et al., 2020). It allows each token in the image to attend to other tokens, effectively considering the relationships between all pixels. This mechanism operates in three steps:

1. *Query, Key, and Value*: Each token is denoted by three vectors: *Q*, *K*, and *V*, respectively.
2. *Attention Score:* The attention scores between each pair of tokens are calculated using a dot product between their query and key vectors (Eqn. 4). Here, $d_k$ denotes key vectors dimensionality.

$$Attention\ (Q,K) = softmax\left(\frac{QK^T}{\sqrt{d_k}}\right) \qquad \text{Eqn. 4}$$

3. *Contextual Representation:* The final representation of each token is obtained by aggregating the value vectors of other tokens weighted by their corresponding attention scores (Eqn. 5).

$$Attention\ (Q,K,V) = softmax\left(\frac{QK^T}{\sqrt{d_k}}\right)V \qquad \text{Eqn. 5}$$

However, ViT leverages multi-head self-attention (MHSA), applying multiple attention mechanisms in parallel to enhance representational power and capture diverse image features.

### 3.2.1 Multi-Head Self-Attention (MHSA)

Each self-attention in MHSA is represented as a head, with each head having its own set of query, key, and value vectors, (Table 4) and (Fig. 7). The outputs of these heads are then combined to create a comprehensive representation of the tokens.

**Table 4: MHSA architectural components**

| Component | Details |
|---|---|
| MHSA | $concat(head_1, head_2 head_3, \dots head_h)W^O$<br>Concatenated multiple attention heads to compute MHSA, where $h$ is number of heads,<br>$W^O$ is output weight matrix. |
| Self-attention | $head_i = softmax\left(\frac{Q_i K_i^T}{\sqrt{d_k}}\right)V_i$<br>i-*th* head denotes a contextual self-attention,<br>$d_k$ denotes the dimension of the key vectors |
| Linear transformation *(Q, K, V)* | $Q_i = XW_Q^i, K_i = X_K^i, V_i = XW_V^i$<br>Where *X* is input |

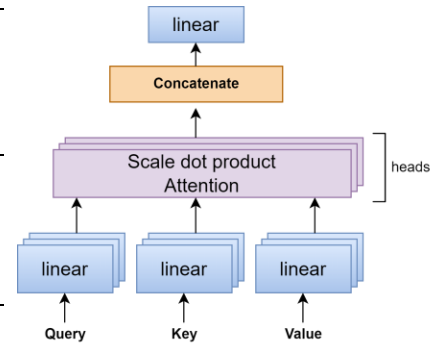

**Fig.7** MHSA

### 3.2.2 Feed-Forward Network (FFN)

The FFN plays a crucial role within ViT, replacing traditional convolutional layers to enable scalable, transformer-based image classification. It processes image patches independently, capturing spatial relationships and extracting key features (Dosovitskiy et al., 2020). This allows ViT to handle images in a tokenized format. To achieve this, the FFN applies a sequence of operations to the output of MHSA layer, introducing non-linearity and boosting the model representational power (Table 5). Given an input sequence $X = \{x_1, x_2, ..., x_n\}$, the FFN in ViT involves a sequence of operations per layer $l$:

**Table 5** FFN architectural components

| Component | Details |
|---|---|
| Skip-connection | $x = x + \tilde{z}^{(l)}$<br>The transformed output $\tilde{z}^{(l)}$ from the MLP block is then added back to the input embeddings through a skip connection, allowing the model to preserve information flow and mitigate vanishing gradient issues. |
| MLP Block | $\tilde{z}^{(l)} = GELU(MLP(z^{(l)}))$<br>The normalized embeddings $z^{(l)}$ are passed through the MLP block, where the internal transformation involves linear projections followed by a GELU activation function. This function contributes to increased non-linearity in the model's transformations, resulting in the extraction of complex features. |
| Layer-normalization | $z^{(l)} = LayerNorm(x)$<br>Input embeddings are initially subjected to layer normalization, aiding in stabilizing and normalizing the activations. |

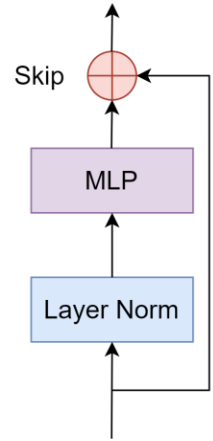

**Fig. 8** FFN

### 3.2.3 Patch-embedding: *Tokenizing image patches*

Unlike CNNs, which process images in a grid-like manner, ViTs divide images into a fixed set of non-overlapping patches and treat each patch as a token (Dosovitskiy et al., 2020). This approach allows ViTs to leverage the power of MHSA in Transformers, enabling them to capture long-range dependencies between pixels and achieve SOTA performance in various image recognition tasks (Lin et al., 2022; Parmar et al., 2018; Liu et al., 2021). Before applying the MHSA mechanism, images are first converted into a sequence of tokens. This is typically done by dividing the image into a grid of patches and representing each patch as a sequence of tokens $P = \{p_1, p_2, ..., p_N\}$, where each $p_i$ denotes a tokenized patch embedding. Let the input image be $X \in \mathbb{R}^{H \times W \times C}$, then patch size is $P \times P$, where $N = \frac{H \times W}{P^2}$ patches (Fig. 9).

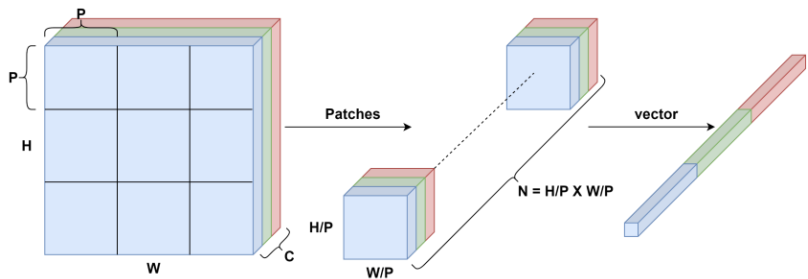

**Fig. 9** Patch-embeddings

### 3.2.4 Positional Encoding: *Preserving Spatial Relationships*

The transformer encoder, being order-agnostic, requires positional encoding for the embedded tokens (Dosovitskiy et al., 2020). This gives them relative spatial information from the original image, aiding the transformer in learning spatial relationships. While there are other types of positional encodings, such as, relative positional embedding (Wu et al., 2021b) and convolutional positional embedding (Wu et al., 2021b). ViT employs an absolute learnable positional embedding (LPE) (Dosovitskiy et al., 2020).

- **Learned Positional Embeddings (LPE)**

LPE capture positional information using a learnable matrix added to initial patch embeddings (Dosovitskiy et al., 2020). To compute LPE, they added a learnable matrix $W_{pos}$, to the initial patch embeddings $P$ described in (Eqn. 6). LPE is effective in preserving spatial information but with increase in computational cost and vary in effectiveness based on tasks and datasets (Dosovitskiy et al., 2020; Wu et al., 2021b).

$$LPE = P + W_{pos} \qquad \text{Eqn. 6}$$

### 3.2.5 ViT transformer block

It's the basic building block of ViTs, consisting of two main components: FFN and MHSA including a series of operations to effectively transform input image sequences into meaningful representations suitable for CV tasks (Table 6), (Fig 10).

Table 6 ViT transformer encoder

| Transformer Blocks in ViTs |
|---|
| **Preprocess input** It starts off with patch embedding, where the input image is divided into non-overlapping patches and each patch is projected onto a lower-dimensional embedding space. Positional encoding is then introduced to supplement the embedded vectors with positional information, preserving the relative positional arrangement of the patches in the input image. |
| **Layer Normalization** Stabilize the internal representation and enhance the learning process, to both the input and output of each MHSA. |
| **MHSA** Is the core component of the Transformer block which enables each token to attend to other tokens in the sequence, capturing long-range dependencies and improves input image global understanding. |
| **MLP** Introduce non-linearity using GELU and expand the expressive power of the model to the output of MHSA. |
| **Skip connections** These connections are integrated around each transformer block. This enables the network to preserve valuable information during training. |
| **Conclude** The interplay of these components within the Transformer block culminates in a powerful and parallelizable architecture that revolutionizes sequence modeling for image recognition tasks. |

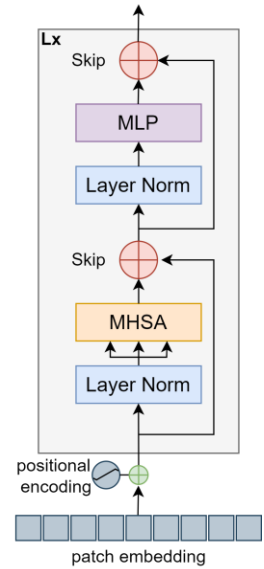

Fig. 10 Transformer block

### 3.2.6 Paradigm Shift in CV

ViT-based models have emerged as paradigm-shifting approaches to solving CV tasks, achieving SOTA performances in image classification and detection, surpassing CNN models. However, ViTs also face limitations, particularly in their high computational cost and potential inefficiencies in handling spatial data. This has led to increasing interest in hybrid models that leverage the synergy of both architectures.

## 4 Hybrid Architectures

Recently, hybrid architectures have emerged as a promising approach to overcome the limitations of both CNNs and ViTs by synergizing their strengths, leading to improved performance in various CV tasks (Peng et al., 2021; Dai et al., 2021; Xu et al., 2021; Carion et al., 2020; Li et al., 2022b; Fang et al., 2022). By understanding these limitations (Table 7) and strengths (Table 8), researchers have developed strategies to mitigate and design hybrid architectures.

**Table 7** CNN and ViT Limitations

| CNN limitations | ViT limitations |
|---|---|
| **Capturing long-range dependencies** CNNs rely on local receptive fields, which limits their ability to capture long-range dependencies between pixels in images. This limits the performance of OD and image segmentation tasks, where understanding the context of distant objects is crucial (Linsley et al., 2018; Yang et al., 2021; Xu et al., 2021). | **Weak local feature extraction** ViTs may struggle with tasks that require strong local feature extraction, such as image recognition tasks with fine-grained details. This is because ViTs focus on capturing global context rather than extracting fine-grained local features (Deininger et al., 2022). |
| **Inductive bias** CNNs exhibit strong inductive bias, which leads to overfitting and poor generalization. This is because CNNs learn local patterns in the training data, which may not generalize well to unseen data (d'Ascoli et al., 2021). | **Sensitivity to noise** ViTs can be sensitive to noisy data, as they rely on global context. Noisy data can poorly affect the self-attention mechanism and lead to poor performance (Bello, 2021). |
| **Fixed Input Size** CNNs typically require fixed input size, making them less adaptable when dealing with images of varying resolutions (Simonyan & Zisserman, 2014b). | **High memory cost** ViTs often involve high computational costs, especially with larger image sizes or more tokens, due to the quadratic complexity of self-attention mechanisms (Zhai et al., 2022). |
| **Global context** the hierarchical structure of CNN limits its ability to capture the entire global context or semantics of an image (Raghu et al., 2021). | **Tokenization Overheads** Dividing images into tokens (patches) might lose fine-grained spatial information, impacting the model's ability to handle intricate local details efficiently (Dosovitskiy et al., 2020). |

**Table 8** Strengths of CNNs and ViTs

| CNN strength | ViT strength |
|---|---|
| **Local Feature Extraction** CNNs excel at extracting and processing local features, making them particularly effective in tasks that require fine-grained details, such as image classification and recognition (Strudel et al., 2021). | **Global Context Modeling** ViTs employ the self-attention mechanism, enabling them to capture long-range dependencies and global contextual relationships between pixels, which is crucial for tasks like object detection and image segmentation (Zhang et al., 2019) |
| **Inductive Biases** this strong bias is due the local connectivity and convolution operations, which improves generalization performance and reduce sensitivity to noisy data (Zhou et al., 2022) | **Scalability** ViTs' image tokenization allows adaptability to varied image sizes, offering versatility within model-defined token limits (Dosovitskiy et al., 2020). |

## 4.1 Hybrid: *Merging CNN and ViT*

Several motivations have driven the design of hybrid models to significantly improve upon specific limitations. We classify this design into parallel, serial, hierarchical architectural integration, early-stage and late-stage feature fusion, and finally, attention mechanism integration (Fig 11).

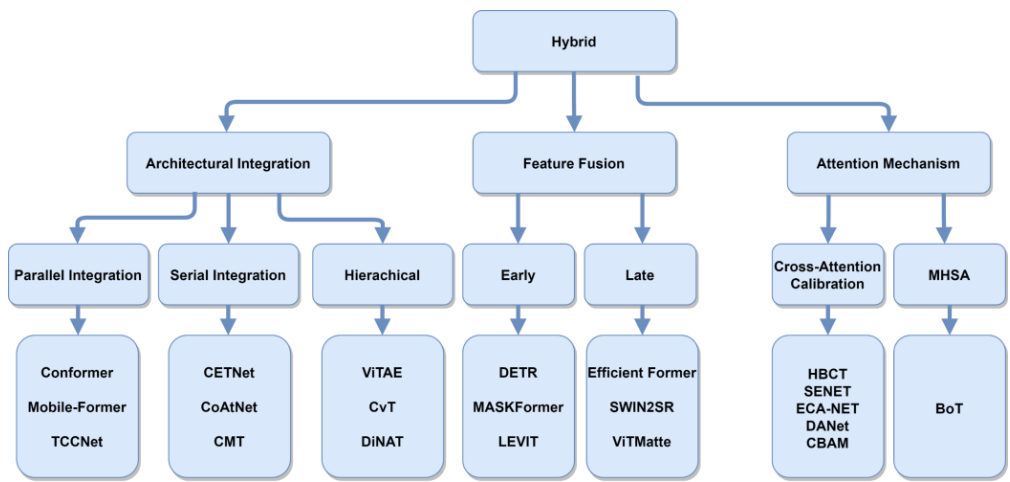

**Fig. 11** Hybrid architectures taxonomy

## 4.2 Architectural Integration: *Parallel Integration*

**Overview** The parallel integration of CNNs and ViTs aims to leverage their individual strengths simultaneously, fostering synergy that allows the model to benefit from both local feature extraction and capturing global context throughout the processing pipeline. In this approach, both CNNs and ViTs operate independently on the input data. They process the information concurrently without sequential dependency. At specific points or layers in the model, the representations extracted by CNNs and ViTs are merged. This fusion can occur through techniques like concatenation, attention mechanisms, or other fusion strategies (Peng et al., 2021; Chen et al., 2022; Li et al., 2022a), (Fig. 12). However, optimizing two distinct architectures running simultaneously is challenging because of its inherent complexity, which also incurs higher computational costs compared to a single monolithic architecture.

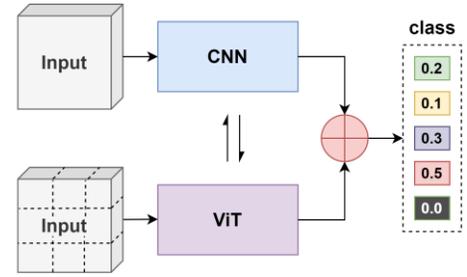

**Fig. 12** Parallel Integration

Peng et al. (2021) introduced the ***Conformer***, a novel parallel hybrid architecture that combines fine-grained local details extracted by CNNs with broader contextual information captured by ViTs. Their innovative Feature Coupling Unit (FCU) aligns and fuses bidirectional features using $1 \times 1$ convolutions for channel alignment and spatial alignment via interpolation (Fig. 13). This unique approach effectively merges local and global representations, resulting in SOTA performance across various image tasks, surpassing both pure CNNs (ResNet, RegNet, DeiT) and ViTs (ViT, T2T, DeiT). Conformer's remarkable performance highlights the synergistic benefits of hybrid architectures in enhancing image analysis. However, its hybrid nature and incorporation of the FCU, alongside parallel processing, inevitably lead to increased computational demands and potential interpretability issues.

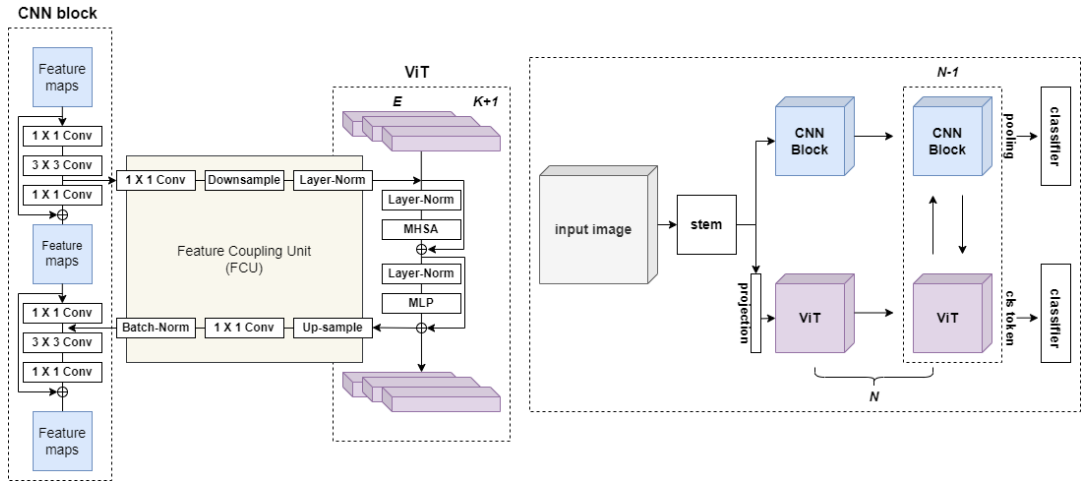

**Fig. 13** Conformer parallel integration of hybrid CNN and ViT

Chen et al., (2022) introduces **Mobile-Former** a novel hybrid mobile vision architecture that combines the efficiency of MobileNets with a lightweight cross-attention mechanism, optimizing computation. This architecture seamlessly integrates local features (former sub-block) and global contextual information (mobile sub-block) through a two-way bridge parallel design (Fig. 14). The *"Mobile to Former"* bridge employs lightweight cross-attention $\mathcal{A}_{X \rightarrow Z}$ to merge local features $X$ with global tokens $Z$, optimizing computation by excluding projection matrices for key and value. Conversely, *"Former to Mobile"* integrates global tokens as key and value with local features as a query $\mathcal{A}_{Z \rightarrow X}$. This innovative design surpasses SOTA CNNs like MobileNetV3, ShuffleNetV2, and EfficientNet-B0, as well as ViTs like T2T-ViT-7, DeiT-Tiny, and Swin-1G in image classification tasks. While demonstrating remarkable synergy in hybrid mobile vision, this architecture faces challenges such as slower inference and reduced accuracy with smaller images due to component inefficiencies. Additionally, it exhibits computational inefficiency in the 'Former' and two-way bridge components for object detection.

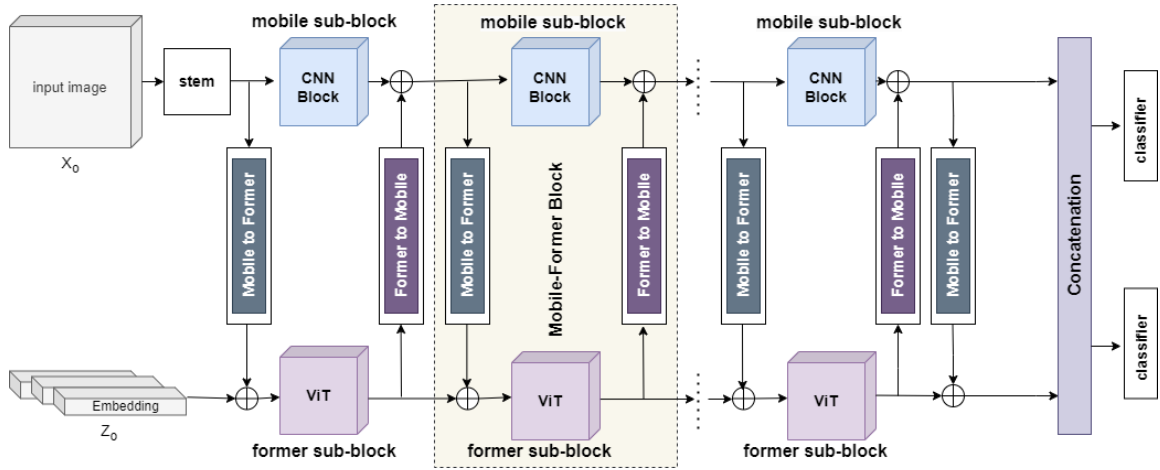

**Fig. 14** Mobile-Former parallel integration of hybrid CNN and ViT

Li et al., (2022a) presents **TCCNet**, a hybrid model designed for person re-identification. This model employs a two-branch parallel architecture, integrating both a CNN branch and a vision transformer branch interconnected by two bridging modules: the Low-level Feature Coupling Module (LFCM) and the High-level Feature Coupling Module (HFCM) (Fig. 15). The LFCM enhances the CNN branch's ability to capture global information by receiving input features from the CNN pathway, conducting operations such as a 1×1 convolution for channel alignment, average pooling, flattening, transpose, and layer normalization before merging with the ViT pathway. In contrast, the HFCM processes input features from the ViT pathway by applying operations like transpose, reshaping to 2D, $1 \times 1$ convolution, batch normalization, and interpolation for spatial alignment, before merging with the CNN pathway. The paper introduces a novel loss function, the duplicate loss, incentivizing each branch to focus on its preferred features. Evaluation on two public datasets, Market1501 (Zheng et al., 2015) and MSMT17 (Wei et al., 2018), showcases the method's improved performance compared to CNN-based (GASM, SPReID, AANet) and ViT-based (DAT, TransReID, PAT, AAformer) models, confirming the synergistic capabilities of hybrid models. While TCCNet demonstrates significant accuracy improvements, its intricate interaction between CNN and ViT components raises challenges in understanding its decision-making processes, posing concerns regarding interpretability.

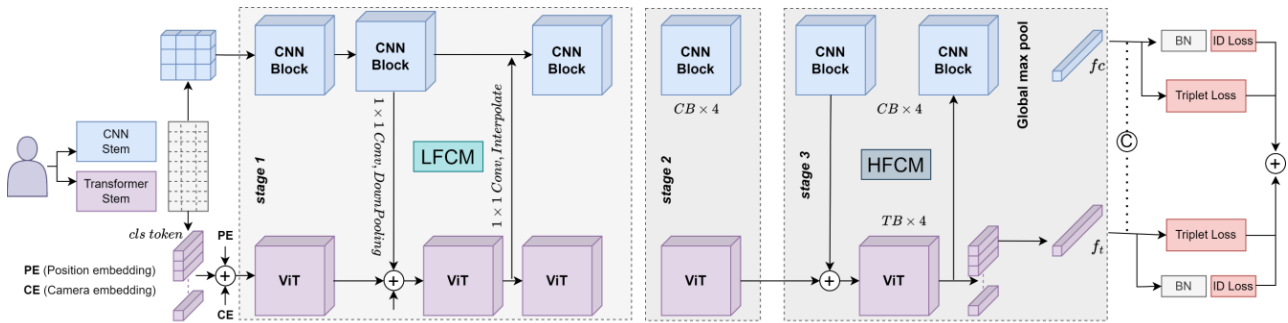

**Fig. 15** TCCNet parallel integration of hybrid CNN and ViT

### 4.3 Architectural Integration: *Sequential Integration*

**Overview** Sequential processing of CNNs and ViTs involves a sequential flow of data, with one architecture initially processing the input and passing its output to the next architecture (Fig. 16). Typically, a CNN extracts local features first, which are then processed by the ViT to capture long-range dependencies. Alignment of representations between architectures is crucial, often necessitating some form of alignment or adaptation, such as resizing, reshaping, or transforming output features from one architecture to match the input format required by the subsequent architecture in sequential processing (Wang et al., 2022; Dai et al., 2021; Guo et al., 2021). However, managing this integration demands careful handling to prevent information loss or distortion caused by mismatches in representations during the sequential transitional phase.

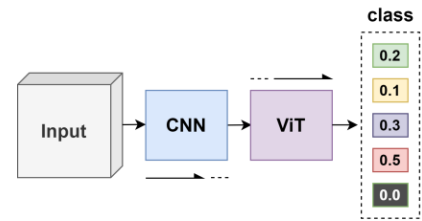

**Fig. 16** Serial Integration

Wang et al. (2022) introduces a novel sequential hybrid approach called *CETNet*, which enhances ViTs by integrating CNNs using Convolutional Embedding (CE) and Locally Enhanced Window Self-Attention (LEWin) mechanisms (Fig. 17). The CE block extracts diverse deep feature representations and injects inductive bias into subsequent ViT attention modules. Meanwhile, LEWin efficiently incorporates local inductive bias via depth-wise separable convolution without significantly increasing computational complexity. This approach utilizes a $3 \times 3$ kernel with stride 1 and padding 1 for convolutional projection on input tokens within a local shift window, involving sequential operations of flatten, Conv2D, Reshape, WHMSA, MLP, and LN. CETNet demonstrates SOTA performance in ImageNet-1k image classification, outperforming ResNet-50, RegNetY-4G, DeiT-S, PVT, and Swin-T, among others. Furthermore, it achieves strong results in segmentation and detection tasks using Mask R-CNN. This hybrid design illustrates the potential synergy between convolutional and transformer-based models, advancing the understanding of effective hybrid model design for CV tasks. However, this design might have limitations in adapting to non-grid structured data and could be sensitive to parameter tuning due to the sequential nature of operations within CE and LEWin.

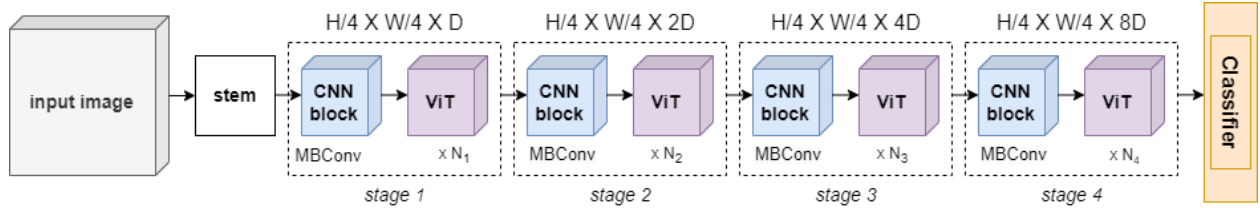

**Fig. 17** CETNet serial integration

Dai et al., 2021 introduces *CoAtNet*, a novel hybrid architecture that merges the strengths of CNNs and ViTs to overcome ViTs inherent limited inductive bias and enhance its generalization capabilities. They achieved this by sequentially feeding the output features from multiple MobileNetV2 (MBConv) blocks into a ViT block (Fig. 18). By adopting this hybrid approach, CoAtNet capitalizes on CNNs' local feature extraction capabilities and ViTs' global context awareness, potentially improving generalization with less data compared to ViTs alone. Evaluation on ImageNet-1k, CoAtNet achieves SOTA results, similar to the performance of ViT-B/16 while having significantly less data. However, optimizing the interplay between MBConv and ViT blocks remains a crucial area for further investigation to fine-tune the model's efficiency and performance across diverse tasks.

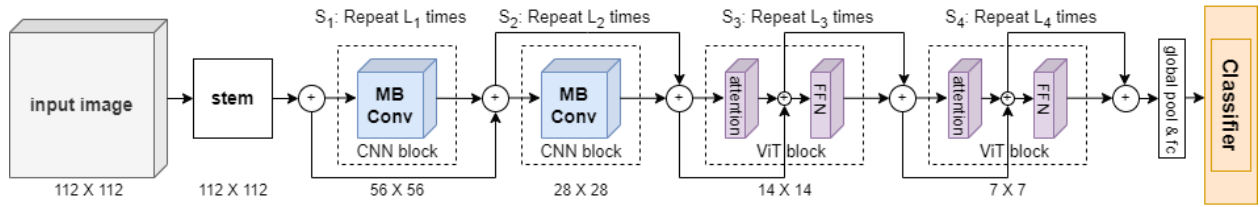

**Fig. 18** CoAtNet serial integration

*CMT,* proposed by Guo et al. (2021), introduces the 'CMT block,' combining $3 \times 3$ depth-wise convolution, lightweight MHSA, and inverted residual FFN. These blocks are sequentially assembled to create the architecture (Fig 19). They harness the strengths of both CNN for efficient local feature extraction and transformers for capturing long-range dependencies, achieving SOTA performance on ImageNet and COCO datasets. The authors show that CMT surpasses standalone CNN and transformer models such as ViT-B16, Swin-B, DenseNet-169, and ResNest50 in accuracy and efficiency. Despite the model's effectiveness in capturing global context, it faces limitations in memory usage.

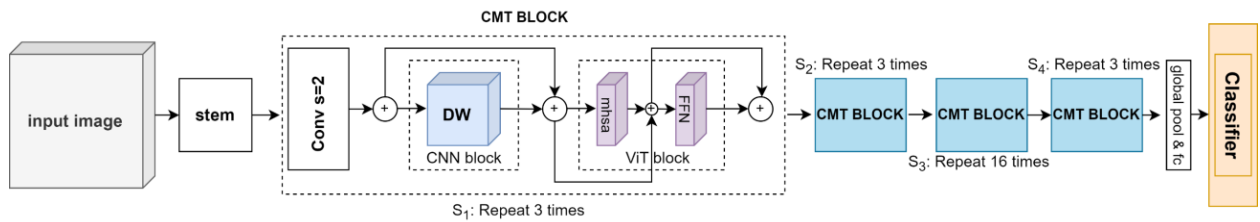

**Fig. 19** CMT serial integration

## 4.4 Architectural Integration: *Hierarchical Integration*

**Overview:** Hierarchical integration merges CNNs and ViTs in a layered approach, leveraging their strengths at different levels of data processing (Fig. 20). CNNs excel in local feature extraction, capturing detailed spatial information, while ViTs specialize in understanding global context and long-range dependencies (Xu et al., 2021; Wu et al., 2021a; Hassani & Shi, 2022). The coherence and alignment of representations between these architectures is key in hierarchical integration. Techniques like cross-layer connections or hierarchical fusion ensure a seamless combination of features across different levels. This careful integration is essential to prevent any information loss across the hierarchical layers, facilitating smooth and effective data transitions between the architectures involved.

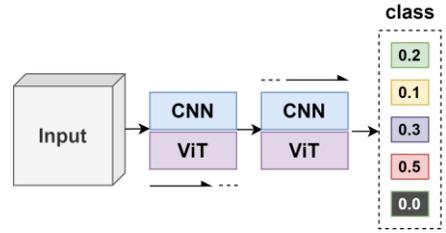

**Fig. 20** Hierarchical Integration

Xu et al. (2021) propose *ViTAE*, a novel vision transformer architecture aimed at enhancing the inductive bias of ViT models to improve their performance across various scales. This is achieved through a hierarchical infusion of intrinsic inductive bias via two key elements: (Fig. 21) (1) Spatial pyramid reduction modules address scale invariance, (2) convolution blocks within transformer layers enable efficient local feature extraction. As a result, they achieve SOTA performance on ImageNet, surpassing benchmarks set by ResNet, DeiT, T2T-ViT-7, Conformer, and other models. This success highlights the benefits derived from hybrid CNN-transformer architectures. However, they face a challenging trade-off in terms of computational efficiency. Their most optimized variant requires a MAC of 20.2, which is relatively high. This high computational demand is a challenge for real-time inference and deployment on resource-constrained devices.

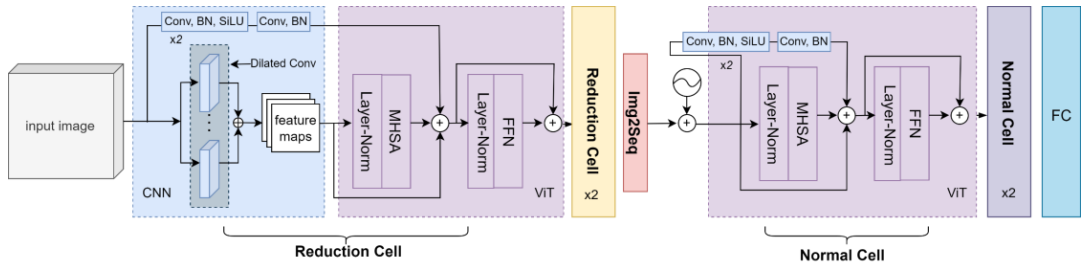

**Fig. 21** ViTAE hierarchical integration

Wu et al. (2021a) propose *CvT* model, which integrates a hierarchical structure that implicitly encodes CNN spatial information, replacing ViT positional encoding. It leverages transformer blocks to process image patches while retaining spatial information crucial for image recognition tasks (Fig. 22). This hybrid model explores the synergy derived from both CNNs' positional encodings and transformers' self-attention mechanisms, resulting in SOTA performance on ImageNet. Their model surpassed ResNet variants and ViT while demonstrating improved parameter efficiency and reduced computational complexity compared to pure transformer models. Despite achieving enhanced accuracy, the significantly higher FLOPS count and number of learnable parameters, in contrast to traditional CNNs, present a challenging trade-off for deployment on constrained devices.

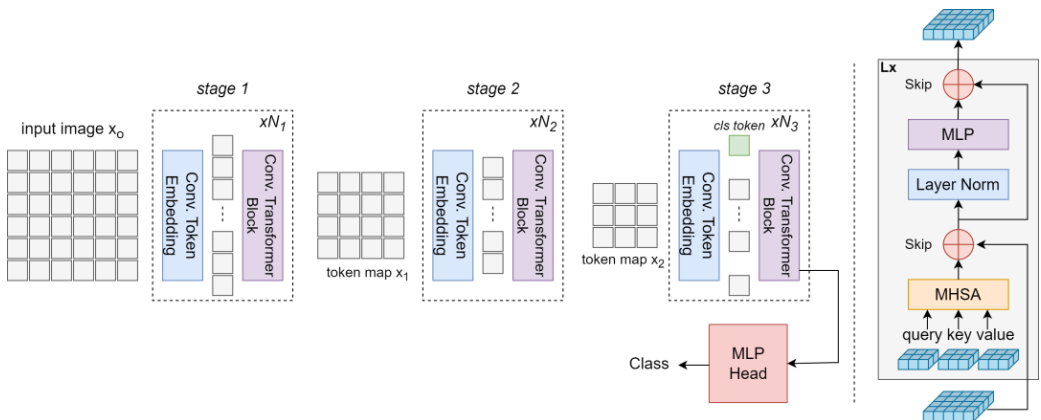

**Fig. 22** CvT hierarchical integration

Hassani & Humphrey (2020) presented **DiNAT**, aimed at tackling the challenges encountered by ViT. DiNAT enhances both computational complexity and global context understanding by introducing Dilated Neighborhood Attention (DNA), a sparse global attention mechanism that efficiently covers the entire image using dilated convolutions. This approach significantly reduces computational complexity compared to ViT's (MHSA) mechanism. DNA is combined with Neighborhood Attention (NA), which focuses on local feature extraction (Fig. 23). This hybrid approach captures comprehensive global context alongside precise local details, allowing DiNAT to excel across various vision tasks. It surpasses Swin and ConvNext variants in object detection, instance segmentation, and semantic segmentation on the COCO benchmark.

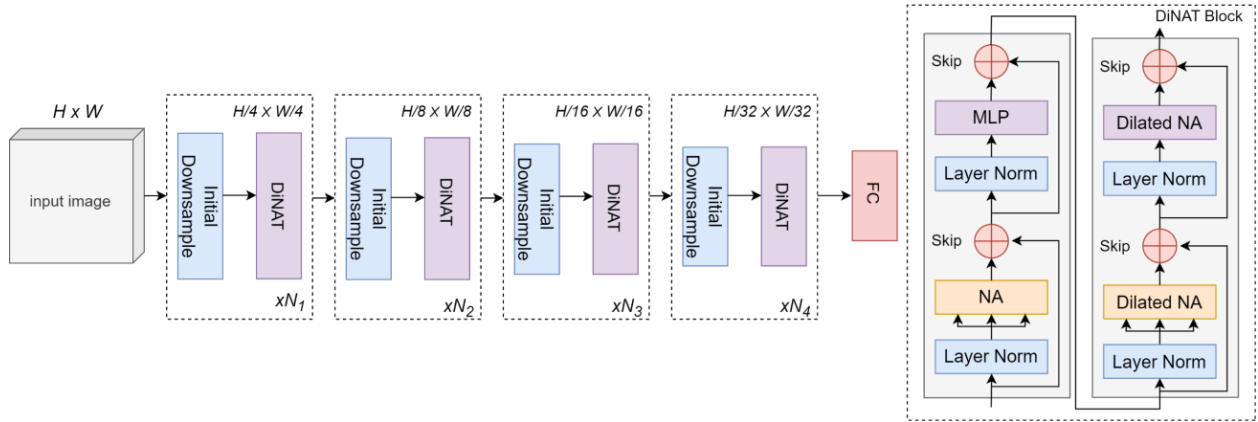

**Fig. 23** DiNAT hierarchical integration

**4.5 Feature Fusion Strategies:** *Early Fusion*

Early fusion in hybrid ViT-CNN architecture aims to leverage the strengths of ViTs in capturing global context and CNNs in local feature extraction from the outset of image processing (Fig. 24). This integration method involves combining the feature maps generated by both models at initial layers before deep hierarchical processing, facilitating simultaneous processing of local details and global relationships (Carion e al., 2020; Cheng et al., 2021; Graham et al., 2021). However, early fusion could lead to a surge in memory and computational cost due to the self-attention mechanism in ViTs, which scales quadratically (Keles et al., 2023). Especially in early stages where features spatial dimension is much larger.

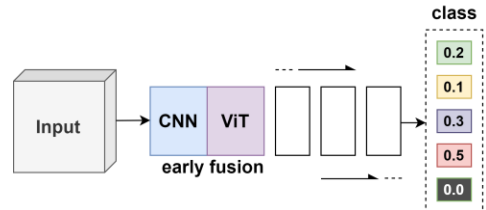

**Fig. 24** Early feature fusion

The **DETR** (Carion et al., 2020) introduces a groundbreaking end-to-end approach to OD by synergizing the capabilities of CNNs and ViTs. This innovative idea transforms the detection paradigm by integrating CNNs in the initial stages to extract image features and ViTs for sequence modeling (Fig. 25). DETR pioneers an end-to-end OD system that eliminates anchor-based methods, showcasing superior accuracy in object localization and recognition compared to Faster-RCNN. However, this pioneering architecture sacrifices some speed when compared to traditional methods. Despite this trade-off, it establishes itself as a transformative architecture that complements CNNs and ViTs for comprehensive OD tasks. Its potential for further optimization in real-time applications remains promising.

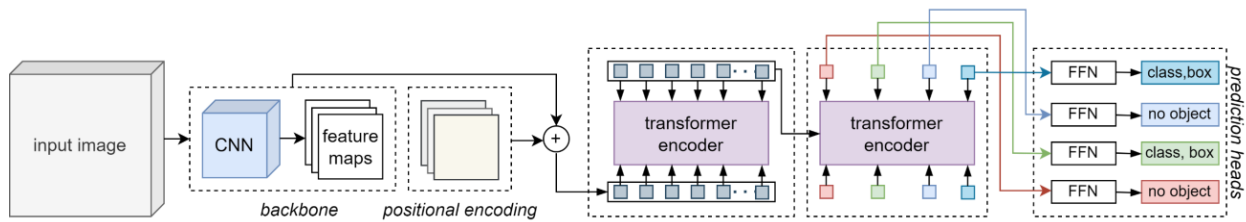

**Fig. 25** DETR early fusion integration

***MASK Former*** (Cheng et al., 2021) introduces a novel approach that unifies semantic and panoptic segmentation through mask classification. The model utilizes a CNN early in the pipeline to extract feature maps. These features are then inputted into a transformer encoder, alongside per-pixel embeddings, to predict a series of binary masks, each linked to a global object category (Fig. 26). This method surpasses per-pixel baseline models, achieving SOTA accuracy, particularly in complex multi-class scenarios such as urban street scenes encompassing various traffic signs and pedestrians. The innovative fusion of CNNs and ViTs holds significant promise for advancements in real-time applications, scene understanding, and domain adaptation within CV. However, there are limitations in detecting new object categories, and challenges in interpretability necessitate further exploration to comprehensively evaluate its potential.

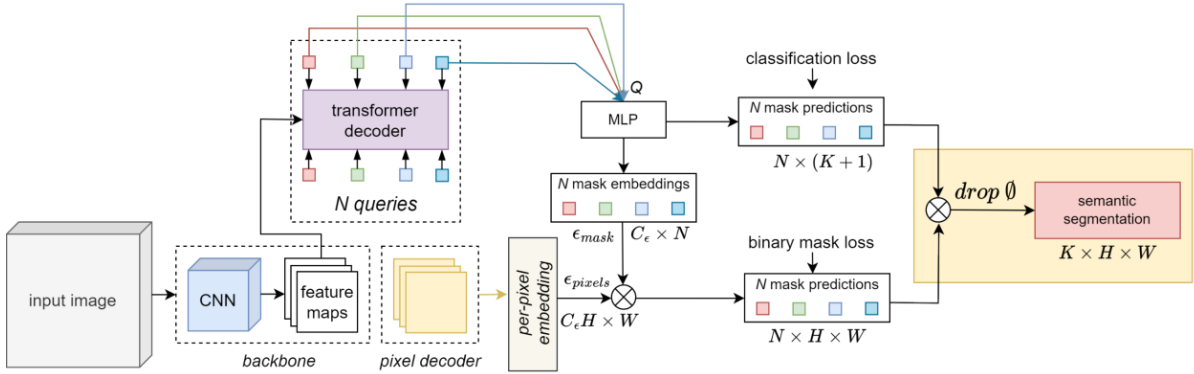

**Fig. 26** MASK Former early fusion integration

***LEVIT*** (Graham et al., 2021) introduces a hybrid approach to image classification that leverages on the synergies between CNNs and ViTs. The model combines the spatially-efficient convolutional stages of CNNs early in the process to reduce spatial dimensions and enhance computational speed, all while retaining crucial spatial information. These features are then integrated into a ViT transformer architecture for final image classification (Fig. 27). This fusion within LeViT models enables flexible speed-accuracy trade-offs, prioritizing computational efficiency while maintaining competitive performance. Notably, LEVIT outperforms Efficient and DeiT variants in Imagenet-1k top-1 accuracy, demonstrating an impressive balance between speed and accuracy.

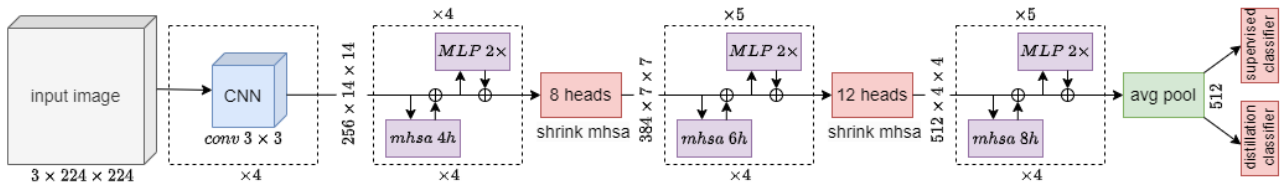

**Fig. 27** LEVIT early fusion integration

**4.6 Feature Fusion Strategies:** *Late Fusion*

Late feature fusion in hybrid ViT-CNN architectures aims to leverage on the strengths of ViTs in capturing global context and CNNs in local feature extraction at later stages of processing (Fig. 28). This method involves independently processing the input data through both models and fusing their extracted representations in deeper layers or at the model's end (Yi et al., 2022; Conde et al., 2022; Yao et al., 2023). Although, merging these distinct representations in later stages might limit the model ability to leverage synergies between local and global information early on, potentially hindering nuanced feature integration.

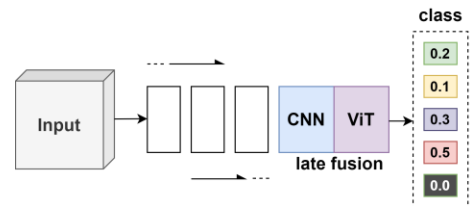

**Fig. 28** Late feature fusion

The ***Efficient Former*** (Yi et al., 2022) is a hybrid model that leverages a convolutional stem and a dimension-consistent pure transformer design specifically optimized for mobile devices. Notably, it achieves this without relying on MobileNet blocks (Fig. 29). This model addresses inefficiencies observed in traditional ViTs by implementing various strategies, such as reducing channel dimensionality and introducing spatial down-sampling stages. It introduces a distinct latency-driven slimming technique that differs from MobileNet approach. Efficient Former demonstrates a significant speed improvement while maintaining competitive accuracy on ImageNet. In fact, it surpasses MobileNet performance on mobile devices in both speed and top-1 accuracy, showcasing its effectiveness in mobile-oriented applications.

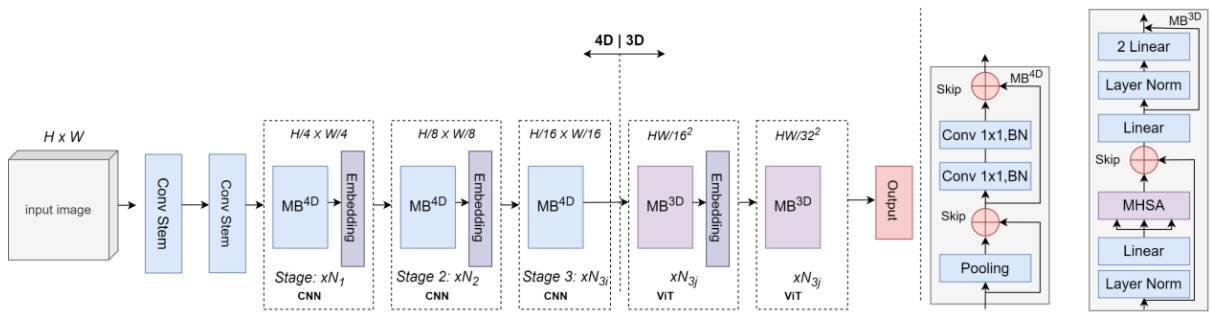

**Fig. 29** Efficient Former late fusion integration

***SWIN2SR*** (Conde et al., 2022) expands the frontiers of image super-resolution by leveraging the strengths of both CNNs and SwinV2 transformers. The model adopts a two-stage approach: initially extracting low-level features using a CNN and subsequently utilizing multiple SwinV2 blocks with advanced attention mechanisms to progressively refine and upscale the feature representation (Fig. 30). This thorough process significantly enhances resolution and detail. The resulting feature maps are then fused to generate the final high-resolution image. This approach outperforms existing SOTA super-resolution methods, including RCAN, SAN, HAN, SwinIR, etc., showcasing notable improvements in both visual quality and quantitative metrics like PSNR and SSIM. However, these advancements come with trade-offs, including increased computational complexity and the potential loss of fine details in complex textures.

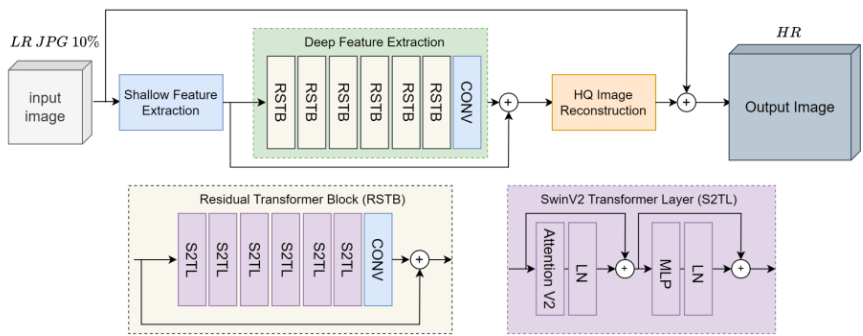

**Fig. 30** SWIN2SR late fusion integration

Yao et al. (2023) proposes ***ViTMatte***, a novel approach to image matting by pioneering the fusion of ViT and CNN. This hybrid architecture harnesses the strengths of both: ViT robust semantic representation capabilities for capturing global context and CNN proficiency in extracting intricate details (Fig. 31). Through a hybrid attention mechanism and a dedicated detail capture module, ViTMatte achieves a superior balance between computational efficiency and precise detail extraction. This synergy propels ViTMatte to SOTA performance on key benchmarks such as Composition1k (Xu et al., 2017) and Distinctions-646 (Qiao et al., 2020), significantly surpassing benchmarks like Closed-Form, KNN, DIM, MatteFomer, Trans Matting, RMat. Results shows the potential of ViT-CNN fusion in advancing image matting tasks and paving way for further exploration.

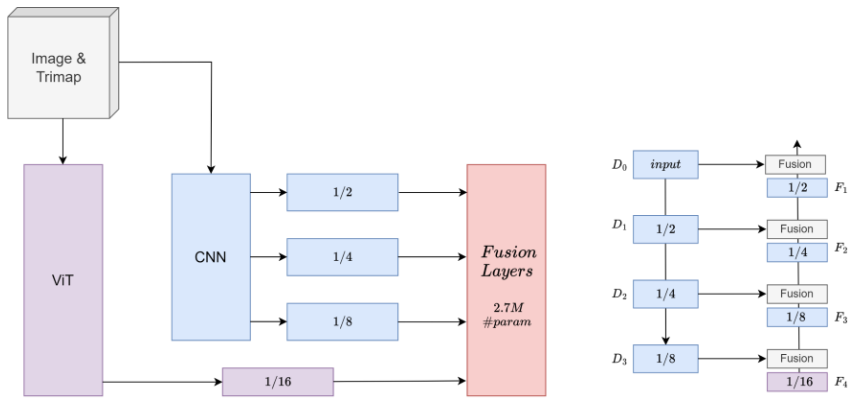

**Fig. 31** ViTMatte late fusion integration

## 4.7 Attention Mechanism Module Integration

***Spatial Attention,*** notably incorporated within CNN architectures, dynamically highlights crucial spatial regions within feature maps (Fig. 32) (Fang et al., 2022). This integration aids tasks like object detection and classification by emphasizing relevant areas while suppressing less informative regions, thereby enhancing network accuracy and efficiency. However, it's predominantly applied within CNNs rather than ViT models for spatially focused tasks.

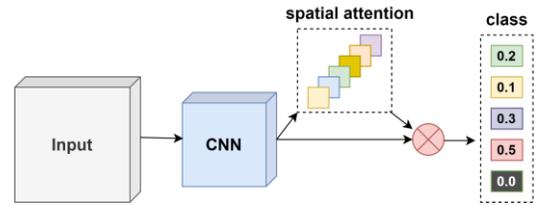

**Fig. 32** Spatial attention calibration

Fang et al. (2022) introduces **HBCT**, a lightweight model for super image resolution. It incorporates various modules, including shallow feature extraction, a hybrid block, dense feature fusion, and an up-sampling module (Fig. 33). Notably, the hybrid block integrates an enhanced spatial attention mechanism at the initial and final stages, enhancing spatial features by emphasizing regions of interest and suppressing irrelevant areas. Ablation studies reveal that removing these spatial attention modules significantly impacts performance. Comparative analysis against SOTA models (SRCNN, FSRCNN, VDSR, DRCN, DRRN, MemNet, IDN, SR-MDNF, CARN, LAPAR-A, IMDN, and RFDN) demonstrates HBCT superior performance, highlighting the effectiveness of incorporating spatial attention. However, this inclusion increases computational complexity, presenting a trade-off between performance gains and computational requirements.

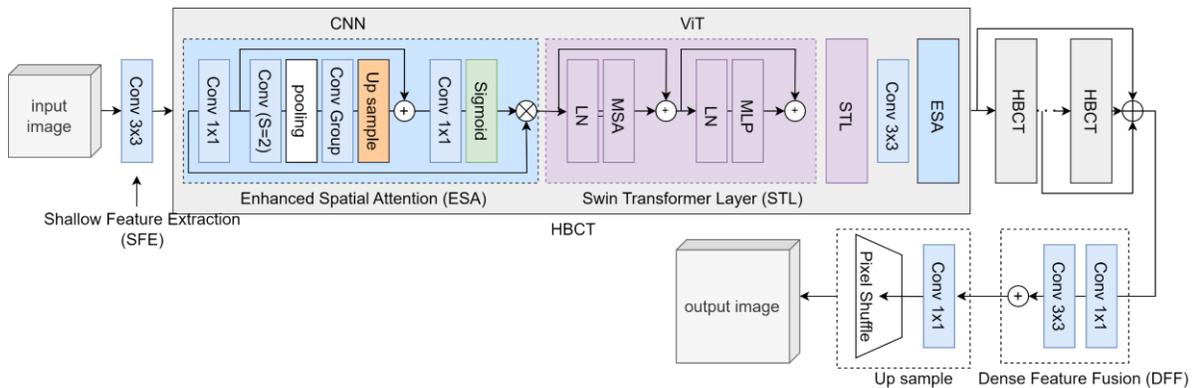

**Fig. 33** HBCT spatial attention integration

***Channel attention*** involves applying attention mechanisms across channels within CNN feature maps (Fig. 34). This technique aids in emphasizing important channels while suppressing less informative ones, thereby enhancing feature selection and channel-wise relevance. By dynamically adjusting channel importance, it helps the network focus on crucial information, contributing to improved feature representation and extraction within the network's architecture (Hu et al., 2018; Wang et al., 2020).

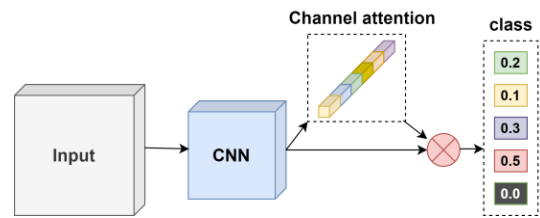

**Fig. 34** Channel attention calibration

Hu et al. (2018) introduced the **SE-NET**, which redefined CNNs by optimizing channel-wise relationships within feature maps through the Squeeze and Excitation (SE) model. This approach involves 'squeeze' (global average pooling) and 'excitation' (dynamic channel recalibration) (Fig. 35), significantly enhancing model performance. Compatible with various architectures such as ResNet-SE, DenseNet-SE, ViT-SE, CvT-SE, LeViT-SE, it effectively improves accuracy and generalization across tasks in CV. However, its implementation might introduce computational overhead. Nonetheless, SE-NET represents a transformative innovation in enhancing various CV tasks.

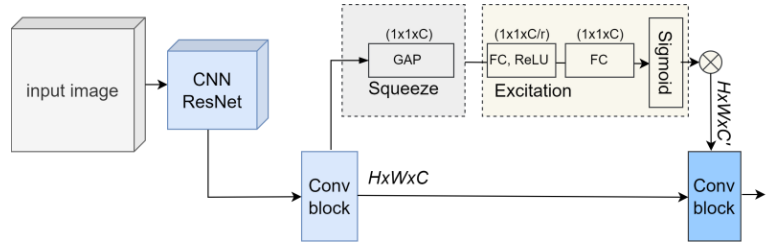

**Fig. 35** SE-NET channel attention calibration

Wang et al. (2020) proposes ***ECA-NET*** which revolutionizes CNN by selectively enhancing informative channels without high computational costs. Unlike SE-NET, ECA uses lightweight 1D convolutional operations along channel dimensions, boosting model accuracy and feature representations (Fig. 36). Its simplicity allows easy integration into standard architectures like ResNet and MobileNet, ViT, providing computational efficiency ideal for resource-limited environments, though its effectiveness may vary based on specific tasks. Overall, ECA stands as an efficient, effective, and lightweight attention mechanism for CNNs.

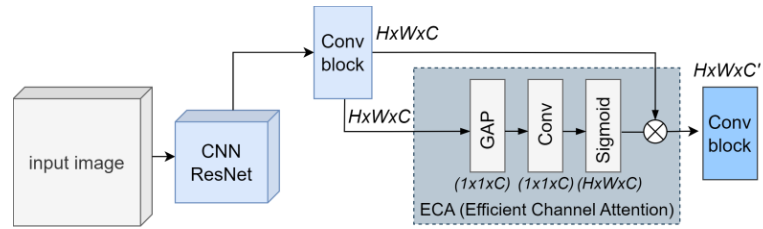

**Fig. 36** ECA-NET channel attention calibration

Combined ***Channel & Spatial Attention*** refers to the application of attention mechanisms across both channels and spatial dimensions within CNN feature maps (Fig. 37). This technique emphasizes important channels while also focusing on specific spatial regions within those channels, aiding in feature selection and enhancing channel-wise relevance. By dynamically adjusting both channel and spatial importance, it helps the network focus on crucial information at both the channel and spatial levels, contributing to improved feature representation and extraction within the network's architecture (Fu et al., 2019; Roy et al., 2018; Woo et al., 2018).

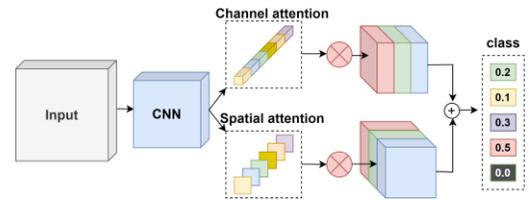

**Fig. 37** channel & spatial attention calibration

Fu et al. (2019) introduced ***DANet***, transforming semantic segmentation by fusing the Positional Attention Module and Channel Attention Module. This dual attention mechanism captures both spatial and channel-wise dependencies concurrently (Fig. 38), significantly enhancing context comprehension. DANet seamlessly integrates into backbone architectures like ResNet, delivering top-tier performance in semantic segmentation across diverse datasets. Yet, its increased computational costs limits deployment in resource-constrained devices. Nonetheless, DANet remains pivotal, excelling in intricate detail capture for precise semantic segmentation tasks.

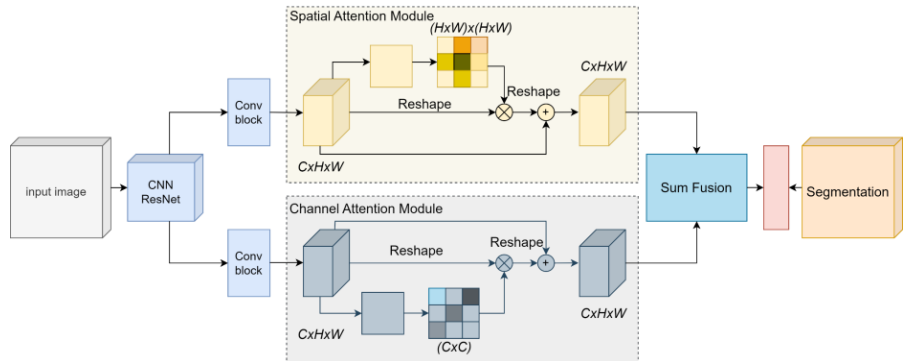

**Fig. 38** DANET combined channel and spatial attention

Roy et al. (2018) introduced *scSE*, enhancing CNN by combining spatial and channel-wise attention mechanisms (Fig. 39). This dual mechanism simultaneously refines spatial features and recalibrates channel responses, significantly enhancing feature representations. scSE easily integrates into prevalent architectures like ResNet and DenseNet, elevating performance in tasks such as classification and segmentation. However, its implementation might lead to increased computational demands. Despite this, scSE remains an effective enhancement for CNNs, empowering feature learning across diverse CV tasks.

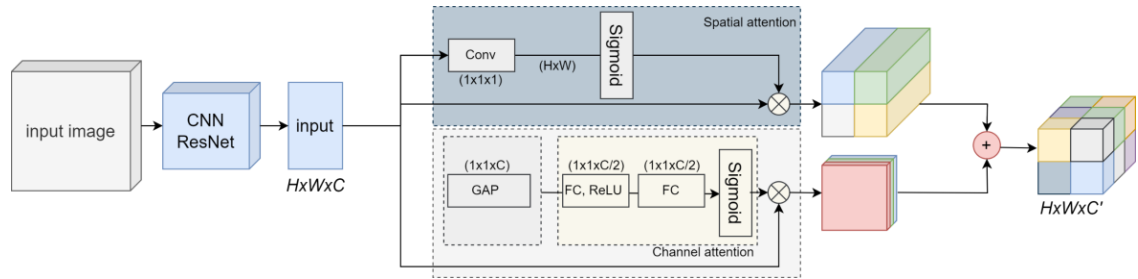

**Fig. 39** scSE combined channel and spatial attention

Woo et al. (2018) introduce *CBAM*, a mechanism that enhances CNNs by integrating spatial and channel-wise attention mechanisms (Fig. 40). This integration refines spatial features and recalibrates channel responses, leading to improved feature representations. Compatible with architectures like ResNet and DenseNet, CBAM significantly enhances performance in tasks such as classification and segmentation. Adapting CBAM for ViT models holds promise in refining feature representations, aiding image understanding tasks. However, its direct integration may necessitate adjustments for ViT self-attention mechanism. Overall, CBAM offers an effective method to elevate feature learning across various CV applications.

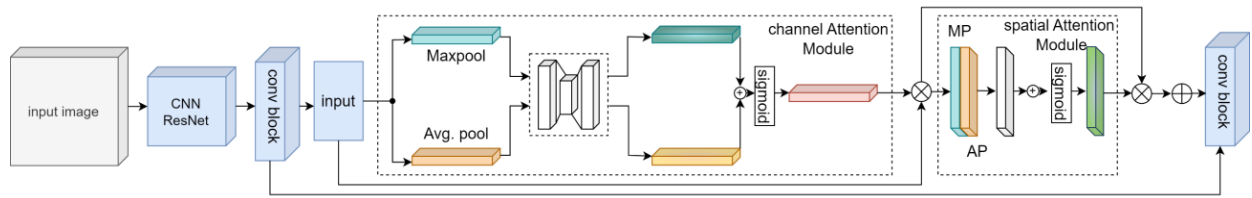

**Fig. 40** CBAM combined channel and spatial attention

*Integrating MHSA into CNN* architectures, involves introducing self-attention layers within the CNN framework (Fig. 41). The MHSA mechanisms enable the model not only to attend to spatially neighboring pixels or regions (as with CNNs) but also to capture long-range dependencies within the data, thereby enhancing the model global context understanding. However, this integration comes with increased computational cost and sensitivity to noisy data (Srinivas et al., 2021).

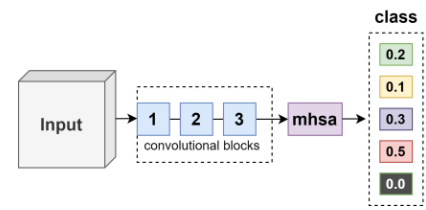

**Fig. 41** MHSA integration in CNN

Srinivas et al. (2021) introduced the *BoTNet* model to enhance object recognition in CV tasks. This model innovatively replaces the standard 3 × 3 convolutional layer in the final block of ResNet with the MHSA mechanism (Fig. 42), a core component of ViT. This strategic integration significantly improves ResNet performance across various tasks. Notably, improves performance on the COCO dataset for instance segmentation, particularly when used with Mask R-CNN. Furthermore, in image classification tasks on ImageNet-1k, BoTNet showcases enhanced performance. A striking observation is that BoTNet self-attention mechanism benefits more from data augmentations like multi-scale jitter compared to ResNet traditional pure convolutions. This finding strongly suggests a successful synergy between ResNet robust feature extraction capabilities and BoTNet MHSA, particularly in capturing long-range dependencies effectively.

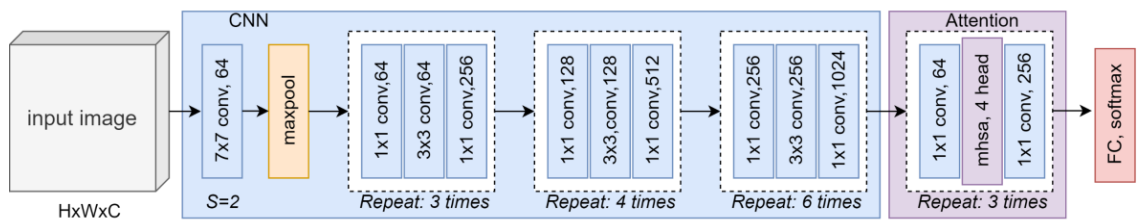

**Fig. 42** BoT MHSA integration

# 5 Comparative Analysis and Applications

This section, performs comparative analysis examining task-specific synergies within hybrid architectures across diverse CV applications. Table 9 shows various models, detailing their parameters, computational resources, and performance metrics across specific tasks, showcasing on their individual strengths and trade-offs. This analysis aims to explain the nuanced performance of these models and highlight their implications in various practical applications.

Table 9 Comparison of Standalone CNN, ViT, and Hybrid Architectures for Task-Specific Applications

| # | CNN | ViT | Hybrid | Task-specific Synergy (application) |
|---|---|---|---|---|
| **Parallel Integration** | | | | |
| Model | ShuffleNetV2 | Swin 2G | Mobile-Former | **Classification Task (IMAGENET1k) @224** |
| Params | 18.1M (+4.1M) | **12.8M** (-1.2M) | 14.0M | *Mobile-Former* (Chen et al., 2022) slightly outperforms Swin 2G in accuracy and parameters, while significantly surpassing ShuffleNetV2 in both accuracy and computational efficiency. |
| MAdds | 573M (+65M) | 2.G (+1.492G) | **508M** | |
| Top-1 | 76.5% (+2.8%) | 79.2% (+0.1%) | **79.3%** | |
| Model | RegNetY32.0GF | DeiT-B | Conformer-B | **Classification Task (IMAGENET1k) @224** |
| Params | 145M (+61.7M) | 86.6M (+3.3M) | **83.3M** | *Conformer-B* (Peng et al., 2021) outperforms RegNetY32.0GF and DeiT-B in accuracy despite having similar parameters to DeiT-B and fewer parameters than RegNetY32.0GF. |
| MACs | 32.3G (+9.0G) | **17.6G** (-5.7G) | 23.3G | |
| Top-1 | 81.0% (+3.1%) | 81.8% (+2.3%) | **84.1%** | |
| Model | ResNet101-BoT | ViT-BoT/s=12 | TCCNet | **Person Re-identification Task (MSMT17)** |
| Params | **42.5M** (-58.7M) | 92.7M (-9.5M) | 101.2 M | *TCCNet* (Li et al., 2022a) notably enhances mAP and Rank-1 scores over ResNet101-BoT & ViT-BoT/s=12, although with higher computational cost. |
| mAP | 57.6% (+9.3%) | 64.4% (+2.5%) | **66.9%** | |
| Rank-1 | 80.1% (+3.6%) | 83.5% (+1.0%) | **84.5%** | |
| **Serial Integration** | | | | |
| Model | RegNetY-16G | Swin-B | CETNet | **Classification Task (IMAGENET1k) @224** |
| Params | 84M (+9M) | 88M (+13M) | **75M** | *CETNet* (Wang et al., 2022) surpasses RegNetY-16G and Swin-B in both computational cost and accuracy. |
| FLOPs | 16.0G (+0.9G) | 15.4G (+0.3G) | **15.1G** | |
| Top-1 | 82.9% (+0.9%) | 83.5% (+0.3) | **83.8%** | |
| Model | NFNet-F5 | CaiT-S-36 | CoAtNet-3 | **Classification Task (IMAGENET1k) @512** |
| Params | 377M (-209M) | **68M** (-100M) | 168M | *CoAtNet-3* (Dai et al., 2021) achieves the same Top-1 accuracy as NFNet-F5 while having fewer parameters and lower computational cost compared to NFNet-F5. |
| FLOPs | 289.8B (+86.7) | **48.B** (-155.1B) | 203.1B | |
| Top-1 | **86.0%** (0%) | 85.0% (+1.0%) | **86.0%** | |
| Model | EfficientNet-B7 | Twins-SVT-L | CMT-L | **Classification Task (IMAGENET1k) @224** |
| Params | **66M** (-8.7M) | 99.2M (+24.5M) | 74.7M | *CMT-L* (Guo et al., 2021) achieves higher top-1 accuracy compared to EfficientNet-B7 & Twins-SVT-L while maintaining a moderate parameter count and computational cost. |
| FLOPs | 37B (+17.5B) | **14.8B** (-4.7B) | 19.5B | |
| Top-1 | 84.3% (+0.5%) | 83.3% (+1.5%) | **84.8%** | |
| **Hierarchical integration** | | | | |
| Model | EfficientNet-B4 | T2T-ViT-14 | ViTAE-S | **Classification Task (IMAGENET1k) @224** |
| Params | **19.3M** (-4.3M) | 21.5M (-2.1M) | 23.6M | *ViTAE-S* (Xu et al., 2021) outperforms both T2T-ViT-14 and EfficientNet-B4 in top-1 accuracy but with higher parameters and computational cost. |
| MACs | 8.4G (-11.8G) | **5.2G** (-15G) | 20.2G | |
| Top-1 | 82.9% (+0.1) | 81.5% (+1.5%) | **83.0%** | |

| Model | ResNet-152 | TNT-B | CvT-W24 | Classification Task (IMAGENET1k) @224 |
|---|---|---|---|---|
| Params | **60M (-217M)** | 66M (-211M) | 277M | *CvT-W24* (Wu et al., 2021a) achieves higher top-1 accuracy though with significantly larger parameters and computational demands in comparisons. |
| FLOPs | **11G (-182.2G)** | 14.1G(-179.1G) | 193.2G | |
| Top-1 | 78.3% (+9.4%) | 82.8% (+4.9%) | **87.7%** | |
| Model | ConvNeXt-B | Swin-B | DiNAT-B | Classification Tasks (IMAGENET1k) @224 |
| Params | 89 M (-1M) | **88 M (-2M)** | 90 M | *DiNAT-B* (Hassani & Humphrey, 2020) achieves improved FLOPs and top-1 accuracy, with parameter counts tradeoff than Swin-B and ConvNeXt-B. |
| FLOPs | 15.4 G (+1.7) | 15.4 G (+1.7) | **13.7 G** | |
| Top-1 | 83.8% (+0.6%) | 83.5% (+0.9%) | **84.4%** | |

| | | **Late Fusion** | | |
|---|---|---|---|---|
| Model | EfficientNet-B5 | CSwin-T | Effc.Former-L7 | Classification Tasks (IMAGENET1k) @224 |
| Params | 30.0M (-52.1M) | **23M (-59.1M)** | 82.1M | *EffcientFormer-L7* (Yi et al., 2022) achieves similar top-1 accuracy with higher parameters and computational cost in comparisons. |
| GMAC | 9.9 (-0.3) | **4.3 (-5.9)** | 10.2 | |
| Top-1 | **83.6% (-0.3%)** | 82.7% (-0.6%) | 83.3% | |
| Model | DnCNN | SwinIR | Swin2SR | JPEG compression artifact reduction Task |
| | *Average* | *Average* | *Average* | (Classic5 datasets) |
| | *PSNR/SSIM* | *PSNR/SSIM* | *PSNR/SSIM* | *Swin2SR* (Conde et al., 2022) outperforms both DnCNN and SwinIR in PSNR/SSIM at $q$ (quality) level of 20, 30 and 40. Although, SwinIR performs better when at $q$ (10). |
| $q$ (10) | 29.40/0.80 | **30.27/0.82** | 30.02/0.81 | |
| $q$ (20) | 31.63/0.86 | 31.32/0.85 | **32.26/0.87** | |
| $q$ (30) | 32.91/0.88 | 31.39/0.853 | **33.51/0.89** | |
| $q$ (40) | 33.77/0.90 | 31.38/0.85 | **34.33/0.90** | |
| Model | RMat | | ViTMatte-B | Image Matting Task (Composition-1k) |
| SAD | 22.87 (+2.54) | | **20.33** | *ViTMatte-B* (Yao et al., 2023) significantly outperforms RMat in comparisons across all metrics. SAD (-2.54), MSE (-0.9), Grad (-1.0) and Conn (-3.06). |
| MSE | 3.9 (+0.9) | *N/A* | **3.0** | |
| Grad | 7.74 (+1.0) | | **6.74** | |
| Conn | 17.84 (+3.06) | | **14.78** | |

| | | **Late Fusion** | | |
|---|---|---|---|---|
| Model | F. RCNN-R101 | | DETR-R101 | Object Detection Task (COCO Dataset) @800x1333 |
| FLOPs | 246G (+7G) | | **253G** | *DETR-R101* (Carion et al., 2020) achieves higher AP scores in comparison to Faster R-CNN-R101 with lower FPS and similar parameters count and FLOPs. |
| FPS | 20 (+10) | *N/A* | **10** | |
| Params | 60M (0) | | **60M** | |
| AP | 44.0 (+0.9) | | **44.9** | |
| Model | DeepLabV3 | Swin-UperNet | MaskFormer | Semantic Segmentation Task (ADE20K) @512 |
| Params | **63M (-149M)** | 234M (-22M) | 212M | *Mask-Former* (Cheng et al., 2021) significantly improves mIoU scores by +9.2 and +2.1 compared to DeepLabV3 and Swin-UperNet. However, with trade-offs in computational cost and fps. |
| FLOPs | **255G (-120G)** | 647G (-272G) | 375G | |
| FPS | **14.2 (-6.3)** | 6.2 (-1.7) | 7.9 | |
| mIoU | 46.4 (+9.2) | 53.5 (+2.1) | **55.6** | |
| Model | EfficientNet-B4 | DeiT-Small | LeViT-384 | Classification Task (IMAGENET1k) @224 |
| Params | **19M (-20.1M)** | 22.5M (-16.6M) | 39.1M | *LeViT-384* (Graham et al., 2021) achieves the same top-1 accuracy as DeiT-Small but with significantly fewer parameters and lower FLOPs. |
| FLOPs | 4200M (+1847M) | 4522M (-2169M) | **2353M** | |
| Top-1 | **82.9% (-0.3%)** | 82.6% (0%) | 82.6% | |

| | Spatial Attention | | | |
|---|---|---|---|---|
| Model | SRCNN | | HBCT | **Super Resolution Task (Urban100 Dataset)** |
| Params | **8K (-364K)** | | 372K | *HBCT* (Fang et al., 2022) outperforms SRCNN in both PSNR and SSIM. Though with higher computational cost than SRCNN. |
| PNSR | 24.52 (+1.68) | *N/A* | **26.20** | |
| SSIM | 0.7221 (+0.0675) | | **0.7896** | |
| | **Channel Attention** | | | |
| Model | ResNet-152 | | ECA-Resnet152 | **Classification Tasks (IMAGENET1k @224)** |
| Params | **57.40M (0.0M)** | | 57.40M | ECA-Resnet152 (Wang et al., 2020) maintains similar parameter and FLOP count as ResNet-152 while achieving +1.34% top-1 accuracy increase. |
| FLOPs | **10.82G (-0.01G)** | *N/A* | 10.83G | |
| Top-1 | 77.58% (+1.34%) | | **78.92** | |
| Model | MobileNet | | MobileNet-SE | **Classification Tasks (IMAGENET1k @224)** |
| Params | **4.2M (-0.5M)** | | 4.7M | *MobileNet-SE* (Hu et al., 2018) shows a slight increase in parameters and FLOPs from MobileNet, yet it achieves a notable +4.1% boost in accuracy. |
| FLOPs | **569M (-3M)** | *N/A* | 572M | |
| Top-1 | 70.6% (+4.1%) | | **74.7%** | |
| | **Combined Channel & Spatial Attention** | | | |
| Model | Dilated FCN | | DANET-Res101 | **Scene Segmentation (Cityscapes Dataset)** |
| basenet | Res101 | *N/A* | Res101 | *DANET* (Fu et al., 2019) significantly increases IoU by +5.03% in comparison to Dilated FCN. |
| IoU | 72.54% (+5.03%) | | **77.57%** | |
| Model | DenseNet | | ScSE-DenseNet | **Segmentation Task (MALC Dataset)** |
| Mean | 0.842 (+0.04) | *N/A* | **0.882** | *ScSE-DenseNet* (Roy et al., 2018) achieves higher mean and std. scores with 0.04 and +±0.005 respectively. |
| std | ± 0.058 (±0.005) | | ± 0.063 | |
| Model | ResNeXt101 | | CBAMResNext | **Classification Tasks (IMAGENET1k @224)** |
| Params | **44.18M (-4.78M)** | | 48.96M | *CBAMResNext* (Woo et al., 2018) achieves top-1 increase in accuracy of 0.47%, slightly outperforming ResNeXt101, with higher computational cost and parameter counts. |
| FLOPs | **7.508G (-0.01G)** | *N/A* | 7.519G | |
| Top-1 | 78.46% (+0.47%) | | **78.93%** | |
| | **MHSA-CNN** | | | |
| Model | Mask-RCNN | | Mask-RCNN | **Instance Segmentation Task (COCO) @512** |
| | ResNet50 | *N/A* | BoT | *BoT* (Srinivas et al., 2021) backbone significantly outperforms ResNet50 with +0.9 increase in $AP^{bb}$ and +0.8 increase in $AP^{mk}$ on Mask-RCNN. |
| $AP^{bb}$ | 42.8 (+0.9) | | **43.7** | |
| $AP^{mk}$ | 37.9 (+0.8) | | **38.7** | |

## 6 Challenges and Future Research Directions

This section outlines key challenges and promising avenues for future research in integrating CNNs and ViTs. Prioritizing the development of efficient and adaptive mechanisms to bridge the feature representations of these architectures could significantly enhance their performance and interpretability. Investigating lightweight ViT variants and exploring novel training algorithms tailored to hybrid architectures are crucial steps towards achieving this goal. Overcoming these challenges has the potential to unlock a new era of DL models with superior performance, interpretability, and generalizability across diverse tasks and domains.

- Aligning the inherently distinct designs of CNNs and ViTs can be challenging. Designing specialized adapters to effectively bridge their feature representations and ensure smooth information flow requires further exploration.
- Matching the representations learned by CNNs and ViTs is crucial. Mismatches can lead to information loss or performance degradation. Future research can focus on developing adaptive mechanisms to dynamically merge their feature spaces.

- Hybrid designs often increase computational complexity due to ViT self-attention mechanism that scales quadratically. Investigating lightweight ViT variants and exploring efficient parallelization techniques are promising avenues for mitigating this challenge.
- Effectively training hybrid models often necessitates substantial resources and time. Balancing the training process to optimize both components while preventing overfitting requires novel optimization algorithms and loss functions tailored to hybrid architectures.
- Combining CNNs' hierarchical representations with ViTs' self-attention mechanisms can affect interpretability. Future research can investigate explainable attention mechanisms and visualization techniques to shed light on how hybrid models make decisions.
- Ensuring good generalization across diverse datasets and effective transfer learning are challenges for hybrid models. Domain adaptation techniques and meta-learning approaches have the potential to improve their adaptability.

## 7 Conclusion

The development of hybrid architectures by combining the strengths of CNNs and ViTs has shown remarkable promise in overcoming the limitations of each individual architecture. By integrating their complementary feature extraction capabilities, hybrid models have achieved significant performance improvements in various CV tasks, such as image classification, object detection, and image segmentation. Our survey of various integration structures, including parallel, serial, hierarchical, and fusion approaches, revealed the synergy between these architectures and their potential to further push the boundaries of DL. However, challenges remain in optimizing hybrid architectures for efficiency, interpretability, and generalizability. Addressing these challenges through lightweight ViT variants, efficient training algorithms, and novel explainable attention mechanisms can unlock the full potential of hybrid models. The future of CV lies in harnessing the combined power of CNNs and ViTs, and hybrid architectures pave the way for a new generation of DL models with superior performance, interpretability, and adaptability across diverse tasks and domains.

**Funding statement**

No funding was received for conducting this study.

**Financial declaration**

The authors have no relevant financial or non-financial interests to disclose.